\documentclass[11pt,a4paper]{article}
\usepackage[hyperref]{emnlp2018}
\usepackage{times}
\usepackage{latexsym}
\usepackage{comment}
\usepackage{url}

\usepackage{wrapfig,lipsum,booktabs}
\usepackage{amsmath}
\usepackage{amssymb}
\usepackage{subfig}
\usepackage{adjustbox} 
\usepackage{caption}

\aclfinalcopy 
\def\aclpaperid{390} 

\title{Learning Context-Sensitive Convolutional Filters for Text Processing}

\author{Dinghan Shen$^{\mathbf{1}}$, ~Martin Renqiang Min$^{\mathbf{2}}$, ~Yitong Li$^{\mathbf{1}}$, ~Lawrence Carin$^{\mathbf{1}}$
	\smallskip
	~  \\
	\smallskip 
	$^{\mathbf{1}}$ Duke University~~~~~~~~~~~~ 
	$^{\mathbf{2}}$ NEC Laboratories America \\
	{\small \tt dinghan.shen@duke.edu, renqiang@nec-labs.com, yitong.li@duke.edu, 
		lcarin@duke.edu} 
	}

\date{}

\begin{document}
\maketitle
\begin{abstract}
Convolutional neural networks (CNNs) have recently emerged as a popular building block for natural language processing (NLP). Despite their success, most existing CNN models employed in NLP share the  same learned (and static) set of filters for all input sentences. In this paper, we consider an approach of using a small \emph{meta network} to learn context-sensitive convolutional filters for text processing. The role of meta network is to abstract the contextual information of a sentence or document into a set of \emph{input-aware} filters. We further generalize this framework to model sentence pairs, where a \emph{bidirectional} filter generation mechanism is introduced to encapsulate co-dependent sentence representations.  In our benchmarks on four different tasks,  including ontology classification, sentiment analysis, answer sentence selection, and paraphrase identification, our proposed model, a modified CNN with context-sensitive filters, consistently outperforms the standard CNN and attention-based CNN baselines. By visualizing the learned \emph{context-sensitive} filters, we further validate and rationalize the effectiveness of proposed framework.



\end{abstract}

\section{Introduction}

In the last few years, convolutional neural networks (CNNs) have demonstrated remarkable progress in various natural language processing applications \cite{collobert2011natural}, including sentence/document classification \citep{kim2014convolutional,zhang2015character, wang2018joint}, text sequence matching \citep{hu2014convolutional,yin2015abcnn, shen2017deconvolutional}, generic text representations \citep{gan2016learning, tang2018speeding}, language modeling \citep{dauphin2016language}, machine translation \citep{gehring2017convolutional} and abstractive sentence summarization \citep{gehring2017convolutional}. 
CNNs are typically applied to tasks where feature extraction and a corresponding supervised task are approached jointly \citep{lecun1998gradient}. 
As an encoder network for text, CNNs typically convolve a set of filters, of window size $n$, with an input-sentence embedding matrix obtained via \emph{word2vec} \citep{mikolov2013distributed} or Glove \citep{pennington2014glove}.
Different filter sizes $n$ may be used within the same model, exploiting meaningful semantic features from different $n$-gram fragments. 

The learned weights of CNN filters, in most cases, are assumed to be fixed regardless of the input text. As a result, the rich contextual information inherent in natural language sequences may not be fully captured. As demonstrated in \citet{cohen1999context},  the context of a word tends to greatly influence its contribution to the final supervised tasks.
This observation is consistent with the following intuition: when reading different {\em types} of documents, \emph{e.g.}, academic papers or newspaper articles, people tend to adopt distinct strategies for better and more effective understanding, leveraging the fact that the same words or phrases may have different meaning or imply different things, depending on context. 

Several research efforts have sought to incorporate \emph{contextual information} into CNNs to adaptively extract text representations. One common strategy is the \emph{attention mechanism}, which is typically employed on top of a CNN (or Long Short-Term Memory (LSTM)) layer to guide the extraction of semantic features. For the embedding of a single sentence, \citet{lin2017structured} proposed a self-attentive model that attends to different parts of a sentence and combines them into multiple vector representations. However, their model needs considerably more parameters to achieve performance gains over traditional CNNs.
To match sentence pairs, \citet{yin2015abcnn} introduced an attention-based CNN model, which re-weights the convolution inputs or outputs, to extract interdependent sentence representations.
\citet{wang2016sentence, wang2016compare} explore a \emph{compare and aggregate} framework to directly capture the word-by-word matching between two paired sentences. However, these approaches suffer from the problem of high matching complexity, since a similarity matrix between pairwise words needs to be computed, and thus it is computationally inefficient or even prohibitive when applied to long sentences \citep{mou2015natural}.

In this paper, we propose a generic approach to learn \emph{context-sensitive} convolutional filters for natural language understanding. In contrast to traditional CNNs, the convolution operation in our framework does not have a fixed set of filters, and thus provides the network with stronger modeling flexibility and capacity. Specifically, we introduce a \emph{meta network} to generate a set of context-sensitive filters, conditioned on specific input sentences; these filters are adaptively applied to either the same (Section~\ref{sec:context}) or different (Section~\ref{sec:adaqa}) text sequences. In this manner, the learned filters vary from sentence to sentence and allow for more fine-grained feature abstraction. 

Moreover, since the generated filters in our framework can adapt to different conditional information available (labels or paired sentences), they can be naturally generalized to model sentence pairs. In this regard, we propose a novel \emph{bidirectional} filter generation mechanism to allow interactions between sentence pairs while constructing \emph{context-sensitive} representations.

We investigate the effectiveness of our Adaptive Context-sensitive CNN (ACNN) framework on several text processing tasks: ontology classification, sentiment analysis, answer sentence selection and paraphrase identification. We show that the proposed methods consistently outperforms the standard CNN and attention-based CNN baselines. Our work provides a new perspective on how to incorporate contextual information into text representations, which can be combined with more sophisticated structures to achieve even better performance in the future.

\section{Related Work}
Learning deep text representations has attracted much attention recently,  since they can potentially benefit a wide range of NLP applications \citep{collobert2011natural, kim2014convolutional, wang2017topic, Shen2018NASHTE, tang2018multi, zhang2018diffusion}. CNNs have been extensively investigated as the encoder networks of natural language. Our work is in line with previous efforts on improving the adaptivity and flexibility of convolutional neural networks \citep{jeon2017active,de2016dynamic}.
\citet{jeon2017active} proposed to enhance the transformation modeling capacity of CNNs by adaptively learning the filter shapes through backpropagation. 
\citet{de2016dynamic} introduced an architecture to generate the future frames conditioned on given image(s), by adapting the CNN filter weights to the motion within previous video frames. Although CNNs have been widely adopted in a large number of NLP applications, improving the adaptivity of vanilla CNN modules has been considerably less studied. To the best of our knowledge, the work reported in this paper is the first attempt to develop more flexible and adjustable CNN architecture for modeling sentences. 

Our use of a meta network to generate parameters for another network is directly inspired by the recent success of hypernetworks for text-generation tasks \citep{ha2016hypernetworks}, and dynamic parameter-prediction for video-frame generation \citep{de2016dynamic}. In contrast to these works that focus on generation problems, our model is based on context-sensitive CNN filters and is aimed at abstracting more informative and predictive sentence features. Most similar to our work, \citet{liu2017dynamic} designed a meta network to generate compositional functions over tree-structured neural networks for encapsulating sentence features. However, their model is only suitable for encoding individual sentences, while our framework can be readily generalized to capture the interactions between sentence pairs. Moreover, our framework is based on CNN models, which is advantageous due to fewer parameters and highly parallelizable computations relative to sequential-based models. 

\section{Model}
\subsection{Basic CNN for text representations} \label{sec:basic}

The CNN architectures in \citep{kim2014convolutional, collobert2011natural} are typically utilized for extracting sentence representations, by a composition of a convolutional layer and a \emph{max-pooling} operation over all resulting feature maps. Let the words of a sentence of length $T$ (padded where necessary) be ${x}_1$, ${x}_2$, $...$ , ${x}_T$. The sentence can be represented as a matrix $\boldsymbol{X} \in \mathbb{R}^{d \times T}$, where each column represents a $d$-dimensional embedding of the corresponding word.

In the convolutional layer, a set of filters with weights $\boldsymbol{W} \in \mathbb{R}^{K \times h \times d}$ is convolved with every window of $h$ words within the sentence, \emph{i.e.}, $\left\{ {x}_{1:h}, {x}_{2:h+1}, \ldotp \ldotp \ldotp, {x}_{T-h+1:T} \right\} $, where $K$ is the number of output feature maps (and filters). In this manner, feature maps $\boldsymbol{p}$ for these $h$-gram text fragments are generated as:
\begin{align}
\boldsymbol{p}_i = f(\boldsymbol{W} \times x_{i:i+h-1} + b) \,
\label{eq:conv}
\end{align}
where $i=1,2,...,T-h+1$ and $\times$ denotes the convolution operator at the $i$th shift location. 
Parameter $b \in \mathbb{R}^{K}$ is the bias term and $f(\cdot)$ is a non-linear function, implemented as a rectified linear unit (ReLU) in our experiments. 
The output feature maps of the convolutional layer, \emph{i.e.}, $ \boldsymbol{p} \in \mathbb{R}^{K \times (T-h+1)} $ are then passed to the pooling layer, which takes the maximum value in every row of $\boldsymbol{p}$, forming a $K$-dimensional vector, $\boldsymbol{z}$. This operation attempts to keep the most salient feature detected by every filter and discard the information from less fundamental text fragments. Moreover, the \emph{max-over-time} nature of the pooling operation \citep{collobert2011natural} guarantees that the size of the obtained representation is independent of the sentence length.

Note that in basic CNN sentence encoders, filter weights are the same for different inputs, which may be suboptimal for feature extraction \citep{de2016dynamic}, especially in the case where conditional information is available.

\subsection{Learning context-sensitive filters} \label{sec:context}
\begin{figure}
	\centering
	\includegraphics[scale=0.39]{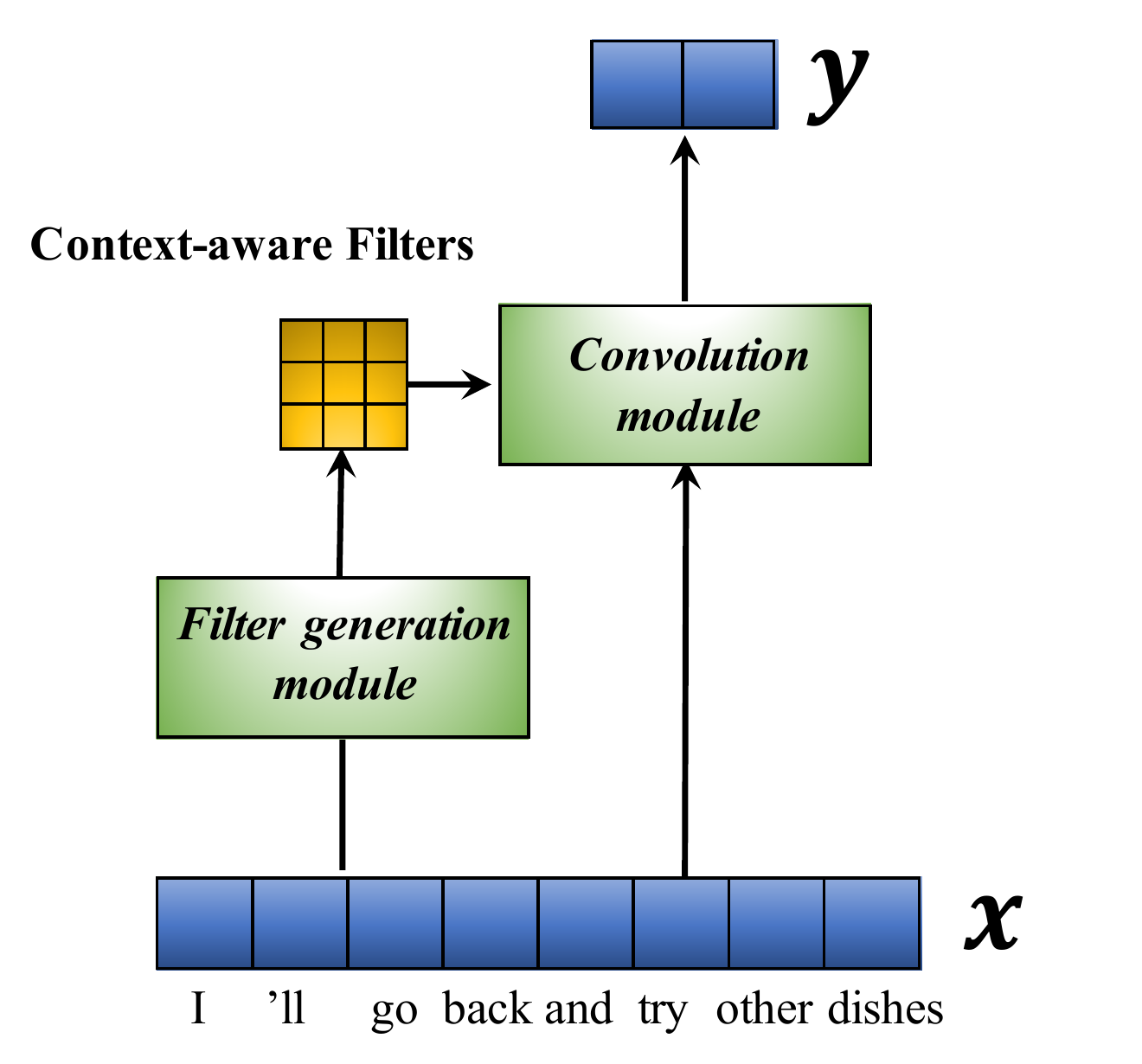}
	\vspace{-3mm}
	\caption{\small The general ACNN framework. Notably, the input sentences to filter generating module and convolution module could be different (see Section~\ref{sec:adaqa}).}
	\label{fig:acnn}
	\vspace{-5mm}
\end{figure}

The proposed architecture to learn context-sensitive filters is composed of two principal modules: (\emph{\romannumeral1}) a filter generation module, which produces a set of filters conditioned on the input sentence; and (\emph{\romannumeral2}) an adaptive convolution module, which applies the generated filters to an input sentence (this sentence may be either the same as or different from the first input, as discussed further in Section~\ref{sec:adaqa}). The two modules are jointly differentiable, and the overall architecture can be trained in an end-to-end manner. Since the generated filters are sample-specific, our ACNN feature extractor for text tends to have stronger predictive power than a basic CNN encoder. The general ACNN framework is shown schematically in Figure~\ref{fig:acnn}.

\paragraph{Filter generation module}
Instead of utilizing fixed filter weights $\boldsymbol{W}$ for different inputs (as \eqref{eq:conv}), our model generates a set of filters conditioned on the input sentence $\boldsymbol{X}$. Given an input $\boldsymbol{X}$, the filter-generation module can be implemented, in principle, as any deep (differentiable) architecture. However, in order to handle input sentences of variable length common in natural language, we design a generic filter generation module to produce filters with a predefined size.
	
First, the input $\boldsymbol{X}$ is encapsulated into a fixed-length vector (code) $\boldsymbol{z}$ with the dimension of $l$, via a basic CNN model, where one convolutional layer is employed along with the pooling operation (as described in Section~\ref{sec:basic}). On top of this hidden representation $\boldsymbol{z}$, a deconvolutional layer, which performs transposed operations of convolutions \cite{radford2015unsupervised}, is further applied to produce a unique set of filters for $\boldsymbol{X}$ (as illustrated in Figure~\ref{fig:acnn}):
\begin{align}
\boldsymbol{z} = & \ \textbf{\normalfont{CNN}}(\boldsymbol{X};  \boldsymbol{\theta}_e),
\label{eq:filter_cnn} \\
\boldsymbol{f} = & \ \textbf{\normalfont{DCNN}}(\boldsymbol{z};  \boldsymbol{\theta}_d) \,, \label{eq:filter_tc}
\end{align}
%
where $\boldsymbol{\theta}_e$ and $\boldsymbol{\theta}_d$ are the learned parameters in each layer of the filter-generating module, respectively. Specifically, we convolve $\boldsymbol{z}$ with a filter of size ($f_s$, $l$, $k_x$, $k_y$), where $f_s$ is the number of generated filters and the kernel size is ($k_x$, ${k_y}$). The output will be a tensor of shape (${f_s}$, ${k_x}$, ${k_y}$). Since the dimension of hidden representation $\boldsymbol{z}$ is independent of input-sentence length, this framework guarantees that the generated filters are of the same shape and size for every sentence.
Intuitively, the encoding part of filter generation module abstracts the information from sentence $\boldsymbol{X}$ into $\boldsymbol{z}$. Based on this representation, the deconvolutional up-sampling layer determines a set of fixed-size, fine-grained filters $\boldsymbol{f}$ for the specific input.

\paragraph{Adaptive convolution module}
The adaptive convolution module takes as inputs the generated filters $\boldsymbol{f}$ and an input sentence. This sentence and the input to the filter-generation module may be identical (as in Figure~\ref{fig:acnn}) or different (as in Figure~\ref{fig:adaqa}). With the sample-specific filters, the input sentence is adaptively encoded, again, via a basic CNN architecture as in Section~\ref{sec:basic}, \emph{i.e.}, one convolutional and one pooling layer. Notably, there are no additional parameters in the adaptive convolution module (no bias term is employed).

Our ACNN framework can be seen as a generalization of the basic CNN, which can be represented as an ACNN by setting the outputs of the filter-generation module to a constant, regardless of the contextual information from input sentence(s). Because of the learning-to-learn \citep{thrun2012learning}
nature of the proposed ACNN framework, it tends to have greater representational power than the basic CNN.

\subsection{Extension to text sequence matching} \label{sec:adaqa} 
Considering the ability of our ACNN framework to generate context-sensitive filters, it can be naturally generalized to the task of text sequence matching. In this section, we will describe the proposed Adaptive Question Answering (AdaQA) model in the context of answer sentence selection task. Note that the corresponding model can be readily adapted to other sentence matching problems  as well (see Section~\ref{sec:exp_qa}). 

Given a factual question $q$ (associated with a list of candidate answers $\left\{  {a_1, a_2, \ldotp \ldotp \ldotp, a_m} \right\}$ and their corresponding labels $\boldsymbol{y}=\left\{  {y_1, y_2, \ldotp \ldotp \ldotp, y_m} \right\}$), the goal of the model is to identify the correct answers from the set of candidates. For $i$ = $1, 2, \ldotp \ldotp \ldotp, m $, if $a_i$ correctly answers $q$, then $y_i = 1$, and otherwise $y_i = 0$. Therefore, the task can be cast as a classification problem where, given an unlabeled question-answer pair $(q_i, a_i)$, we seek to predict the judgement $y_i$.

\begin{figure}[t!]
	\centering
	\includegraphics[scale=0.34]{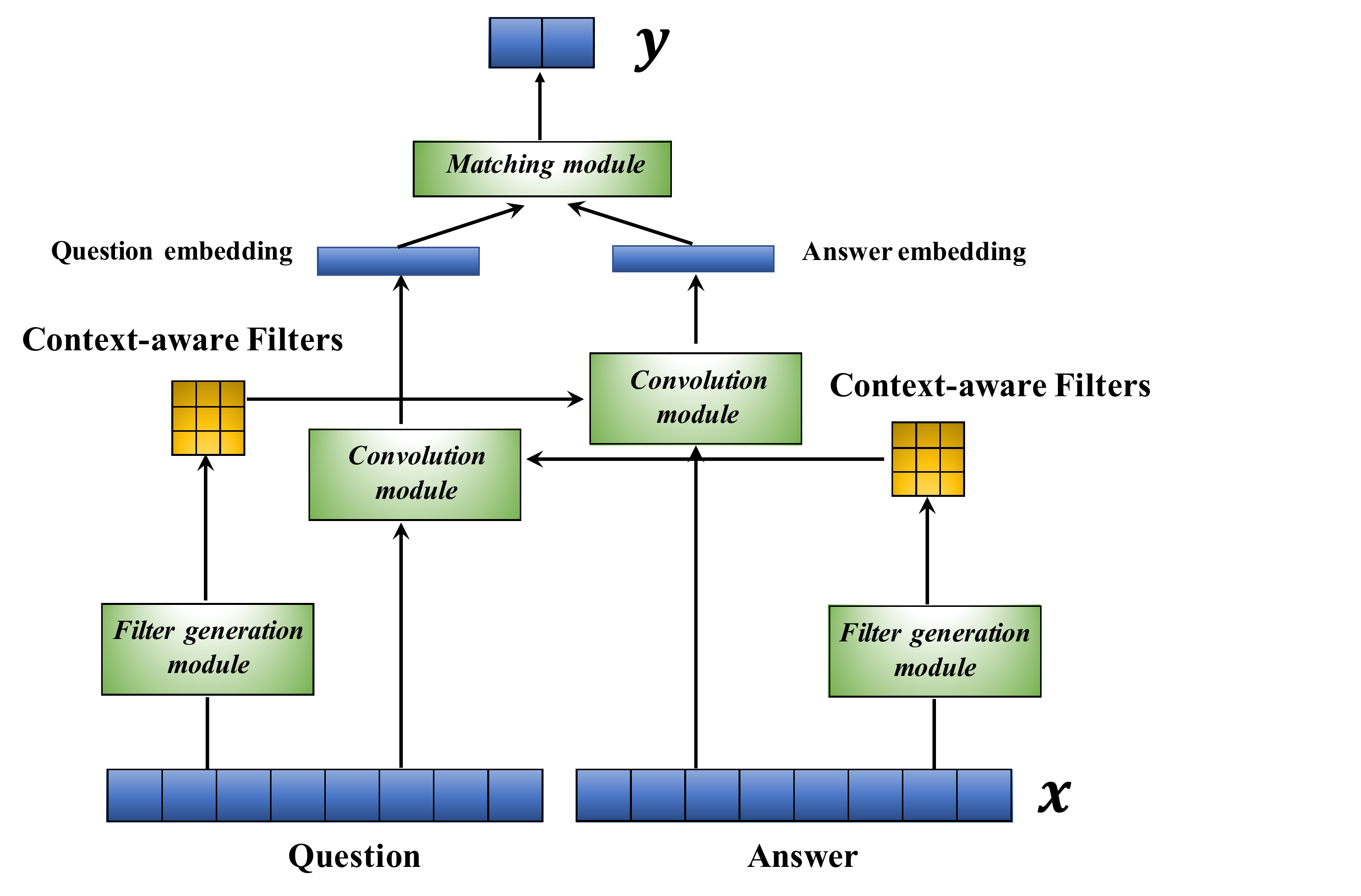}
	\vspace{-3mm}
	\caption{\small Schematic description of Adaptive Question Answering (AdaQA) model. }
	\label{fig:adaqa}
	\vspace{-3mm}
\end{figure}

Conventionally, a question $q$ and an answer $a$ are independently encoded by two basic CNNs to fixed-length vector representations, denoted $\boldsymbol{h}_q$ and $\boldsymbol{h}_a$, respectively. They are then directly employed to predict the judgement $\boldsymbol{y}$. This strategy could be suboptimal, since no communication (information sharing) occurs between the question-answer pair until the top prediction layer. Intuitively, while the model is inferring the representation for a question, if the meaning of the answer is taken into account, those features that are relevant for final prediction are more likely to be extracted. So motivated, we propose an adaptive CNN-based question-answer (AdaQA) model for this problem. 
The AdaQA model can be divided into three modules: filter generation, adaptive convolution, and matching modules, as depicted schematically in Figure~\ref{fig:adaqa}. Assume there is a question-answer pair to be matched, represented by word-embedding matrices, \emph{i.e.} $\boldsymbol{Q} \in \mathbb{R}^{T_q \times d}$ and $\boldsymbol{A} \in \mathbb{R}^{T_a \times d}$, where $d$ is the embedding dimension and $T_q$ and $T_a$ are respective sentence lengths. First, they are passed to two filter-generation modules, to produce two sets of filters that encapsulate features of the corresponding input sentences. Similar to the setup in Section~\ref{sec:context}, we also employ a two-step process to produce the filters. For a question $\boldsymbol{Q}$, the generating process is:
\begin{align}
\vspace{-8mm}
\boldsymbol{z}_q = & \ \textbf{\normalfont{CNN}}(\boldsymbol{Q}; \boldsymbol{\theta}_{e}^q) \label{eq:filter_cnn}, \\
\boldsymbol{f}_q = & \ \textbf{\normalfont{DCNN}}(\boldsymbol{z}_q; \boldsymbol{\theta}_{d}^q) \, \label{eq:filter_dcnn}
\vspace{-8mm}
\end{align}
where CNN and DCNN denote the basic CNN unit and deconvolution layer, respectively, as discussed in Section 2.1. Parameters $\boldsymbol{\theta}_{e}^q$ and $\boldsymbol{\theta}_{d}^q$ are to be learned. The same process can be utilized to produce encodings $\boldsymbol{z}_a$ and filters $\boldsymbol{f}_a$ for the answer input, $\boldsymbol{A}$, with parameters $\boldsymbol{\theta}_{e}^a$ and $\boldsymbol{\theta}_{d}^a$, respectively. 

The two sets of filter weights are then passed to adaptive convolution modules, along with $\boldsymbol{Q}$ and $\boldsymbol{A}$, to obtain the extracted question and answer embeddings.
That is, the question embedding is convolved with the filters produced by the answer and \emph{vise versa} ($\boldsymbol{\psi}_q$ and $\boldsymbol{\psi}_a$ are the bias terms to be learned). The key idea is to abstract information from the answer (or question) that is pertinent to the corresponding question (or answer). Compared to a Siamese CNN architecture \citep{bromley1994signature}, our model selectively encapsulates the most important features for judgement prediction, removing less vital information. We then employ the question and answer representations $\boldsymbol{h}_q \in \mathbb{R}^{n_h}$, $\boldsymbol{h}_a \in \mathbb{R}^{n_h}$ as inputs to the matching module (where $n_h$ is the dimension of question/answer embeddings).
Following \citet{mou2015natural}, the matching function is defined as:
\begin{align}
\boldsymbol{t} = [\boldsymbol{h}_q; \boldsymbol{h}_a; \boldsymbol{h}_q - \boldsymbol{h}_a; \boldsymbol{h}_q \odot \boldsymbol{h}_a] \\
p(\boldsymbol{y}=1 | \boldsymbol{h}_q, \boldsymbol{h}_a) = \textbf{\normalfont{MLP}}(\boldsymbol{t}; \boldsymbol{\eta}^{\prime})
\label{eq:predict}
\end{align}
where $ - $ and $\odot$ denote an element-wise subtraction and element-wise product, respectively. $[\boldsymbol{h}_a ; \boldsymbol{h}_b]$ indicates that $\boldsymbol{h}_a$ and $\boldsymbol{h}_b$ are stacked as column vectors. The resulting matching vector $\boldsymbol{t} \in \mathbb{R}^{4n_h}$ is then sent through an MLP layer (with sigmoid activation function and parameters $\boldsymbol{\eta}^{\prime}$ to be learned) to model the desired conditional distribution $p(y_i=1 | \boldsymbol{h}_q, \boldsymbol{h}_a)$.

Notably, we share the weights of filter generating networks for both the question and answer, so that the model adaptivity for answer selection can be improved without an excessive increase in the number of parameters. All three modules in AdaQA model are jointly trained end-to-end. Note that the AdaQA model proposed can be readily adapted to other sentence matching tasks, such as paraphrase identification (see Section~\ref{sec:exp_qa}).

\subsection{Connections to attention mechanism}
\label{sec:att}
The adaptive \emph{context-sensitive} filter generation mechanism proposed here bears close resemblance to attention mechanism \cite{yin2015abcnn, bahdanau2014neural, xiong2016dynamic} widely adopted in the NLP community, in the sense that both methods intend to incorporate rich \emph{contextual information} into text representations. However, attention is typically operated on top of the hidden units preprocessed by CNN or LSTM layers, and assigns different weights to each unit according to a \emph{context vector}. By contrast, in our \emph{context-sensitive} filter generation mechanism, the contextual information is inherently encoded into the convolutional filters, which directly interact with the input sentence during the convolution encoding operation. 
Notably, according to our experiments, the proposed filter generation module can be readily combined with (standard) attention mechanisms to further enhance the modeling expressiveness of CNN encoder.

\section{Experimental Setup}
\paragraph{Datasets}
We investigate the effectiveness of the proposed ACNN framework on both document classification and text sequence matching tasks. Specifically, we consider two large-scale document classification datasets: Yelp Reviews Polarity, and DBPedia ontology datasets \citep{zhang2015character}. For Yelp reviews, we seek to predict a binary label (positive or negative) regarding one review about a restaurant. DBpedia is extracted from Wikipedia by crowd-sourcing and is categorized into 14 non-overlapping ontology classes, including \emph{Company}, \emph{Athlete}, \emph{Natural Place}, \emph{etc}. We sample 15$\%$ of the training data as the validation set, to select hyperparameters for our models and perform early stopping. 
For sentence matching, we evaluate the AdaQA model on two datasets for open-domain question answering: WikiQA \citep{yang2015wikiqa} and SelQA \citep{jurczyk2016selqa}. Given a question, the task is to rank the corresponding candidate answers, which, in the case of WikiQA, are sentences extracted from the summary section of a related Wikipedia article. 
To facilitate comparison with existing results \citep{yin2015abcnn,yang2015wikiqa, Shen2018BaselineNM}, we truncate the candidate answers to a maximum length of 40 tokens for all experiments on the WikiQA dataset.
We also consider the task of paraphrase identification with the Quora Question Pairs dataset, with the same data splits as in \cite{wang2017bilateral}. 
A summary of all datasets is presented in Table~\ref{tab:summary}.

\begin{table}[t!] \small
	\centering
	\begin{tabular}{cccccc}
		\toprule[1.2pt]
		\textbf{Dataset} & \textbf{\# \textbf{train}/ \textbf{test}} & \textbf{average \#w} & \textbf{vocabulary} \\
		\hline
		Yelp P.       &  560k/ 38k & 138 & 25,709  \\
		DBpedia         & 560k/ 70k & 56 &  21,666 \\
		\hline 
		WikiQA       &  20,360/ 2,352 & 7/ 26 & 10,000  \\
		SelQA         & 66,438/ 19,435 & 8/ 24 &  20,000 \\
		Quora       & 390k/ 5,000 & 13/ 13 & 20,000  \\
		\bottomrule[1.2pt]
	\end{tabular}
	\vspace{-3mm}
	\caption{\small Dataset statistics.}
	\label{tab:summary}
	\vspace{-4mm}
\end{table}

\paragraph{Training Details}
For the document classification experiments, we randomly initialize the word embeddings uniformly within $[-0.001, 0.001]$ and update them during training. For the generated filters, we set the window size as $h=5$, with $K=100$ feature maps (the dimension of $z$ is set as 100). For direct comparison, we employ the same filter shape/size settings as in our basic CNN implementation. A one-layer architecture is utilized for both the CNN baseline and the ACNN model, since we did not observe significant performance gains with a multilayer architecture. 
The minibatch size is set as 128, and a dropout rate of 0.2 is utilized on the embedding layer. We observed that a larger dropout rate (\emph{e.g.}, 0.5) will hurt performance on document classifications and make training significantly slower. 

For the sentence matching tasks, we initialized the word embeddings with 50-dimensional Glove \citep{pennington2014glove} 
word vectors pretrained from Wikipedia 2014 and Gigaword 5 \citep{pennington2014glove} for all model variants. As for the filters, we set the window size as $h=5$, with $K=300$ feature maps. As described in Section~\ref{sec:adaqa}, the vector $\boldsymbol{t}$, output from the matching module, is fed to the prediction layer, implemented as a one-layer MLP followed by the sigmoid function. We use Adam \citep{kingma2014adam} to train the models, with a learning rate of $3 \times 10^{-4}$. Dropout \citep{srivastava2014dropout}, with a rate of 0.5, is employed on the word embedding layer. The hyperparameters are selected by choosing the best model on the validation set. All models are implemented with TensorFlow \cite{abadi2016tensorflow} and are trained using one NVIDIA GeForce GTX TITAN X GPU with 12GB memory. \par
\paragraph{Baselines}
For document classification, we consider several baseline models: 
($i$) ngrams \citep{zhang2015character}, a bag-of-means method based on TFIDF representations built by choosing the 500,000 most frequent n-grams (up to 5-grams) from the training set and use their corresponding counts as features; 
($ii$) small/large word CNN \citep{zhang2015character}: 6 layer word-based convolutional networks, with 256/1024 features at each layer, denoted as small/large, respectively;
($iii$) deep CNN \citep{conneau2016very}: deep convolutional neural networks with 9/17/29 layers.
To evaluate the effectiveness of proposed AdaQA model, we compare it with several CNN-based sequence matching baselines, including Vanilla CNN \citep{jurczyk2016selqa,santos2017learning}, attentive pooling networks \citep{dos2016attentive}, and ABCNN \citep{yin2015abcnn} (where an attention mechanism is employed over the two sentence representations).
\paragraph{Evaluation Metrics}
For document categorization and  paraphrase identification tasks, we employ the percentage of correct predictions on the test set to evaluate and compare different models. For the answer sentence selection task, mean average precision (MAP) and mean reciprocal rank (MRR) are utilized as the corresponding evaluation metrics.

\begin{table}[t!]
	\centering
	\begin{tabular}{c|c|c}
		\toprule[1.2pt]
		\textbf{Model} &  	\textbf{Yelp P.} &  	\textbf{DBpedia}  \\
		\hline
		\multicolumn{3}{c}{\emph{\textbf{CNN-based Baseline Models}}} \\
		\hline
		ngrams$^{\ast}$     & 4.36 &  1.37     \\
		ngrams TFIDF$^{\ast}$     & 4.56 &  1.31     \\
		Small word CNN$^{\ast}$        & 5.54 & 1.85  \\ 
		Large word CNN$^{\ast}$        & 4.89 & 1.72  \\ 
		Self-attentive Embedding $^{\ddagger}$    & 3.92 & 1.14  \\ 
		Deep CNN (9 layer)$^{\dagger}$         &  4.88 & 1.35  \\
		Deep CNN (17 layer)$^{\dagger}$           &  4.50 &  1.40 \\
		Deep CNN (29 layer)$^{\dagger}$       & 4.28 &  1.29 \\
		\hline
		\multicolumn{3}{c}{\emph{\textbf{Our Implementations}}} \\
		\hline
		S-CNN         & 14.48 & 22.35 \\
		S-ACNN         & \textbf{6.41} & \textbf{5.16}  \\
		\hline
		M-CNN         & 4.58 & 1.66 \\
		M-ACNN         & \textbf{3.89} & \textbf{1.07}  \\
		\bottomrule[1.2pt]
	\end{tabular}
	\caption{\small Test error rates on document classification tasks (in percentages). S-\emph{model} indicates that the \emph{model} has one single convolutional filter, while M-\emph{model} indicates that the \emph{model} has multiple convolutional filters. Results marked with $\ast$ are reported by \cite{zhang2015character}, $\dagger$ are reported by \cite{conneau2016very}, and $\ddagger$ are reported by \cite{lin2017structured}.}
	\label{tab:topic}
	\vspace{-4mm}
\end{table}

\section{Experimental Results}
\subsection{Document Classification}
To explicitly explore whether our ACNN model can leverage the input-aware filter weights for better sentence representation, we perform a comparison between the basic CNN and ACNN models with only a \emph{single} filter, which are denoted as S-CNN, S-ACNN, respectively (this setting may not yield best overall performance, since only a single filter is used, but it allows us to isolate the impact of adaptivity). As illustrated in Table~\ref{tab:topic}, S-ACNN significantly outperforms S-CNN on both datasets, demonstrating the advantage of the filter-generation module in our ACNN framework. As a result, with only one convolutional filter and thus very limited modeling capacity, our S-ACNN model tends to be much more expressive than the basic CNN model, due to the flexibility of applying different filters to different sentences.

\begin{figure*} \centering
	\vspace{-2mm}
	\begin{tabular}{ccc}  
		\includegraphics[width=5.0cm]{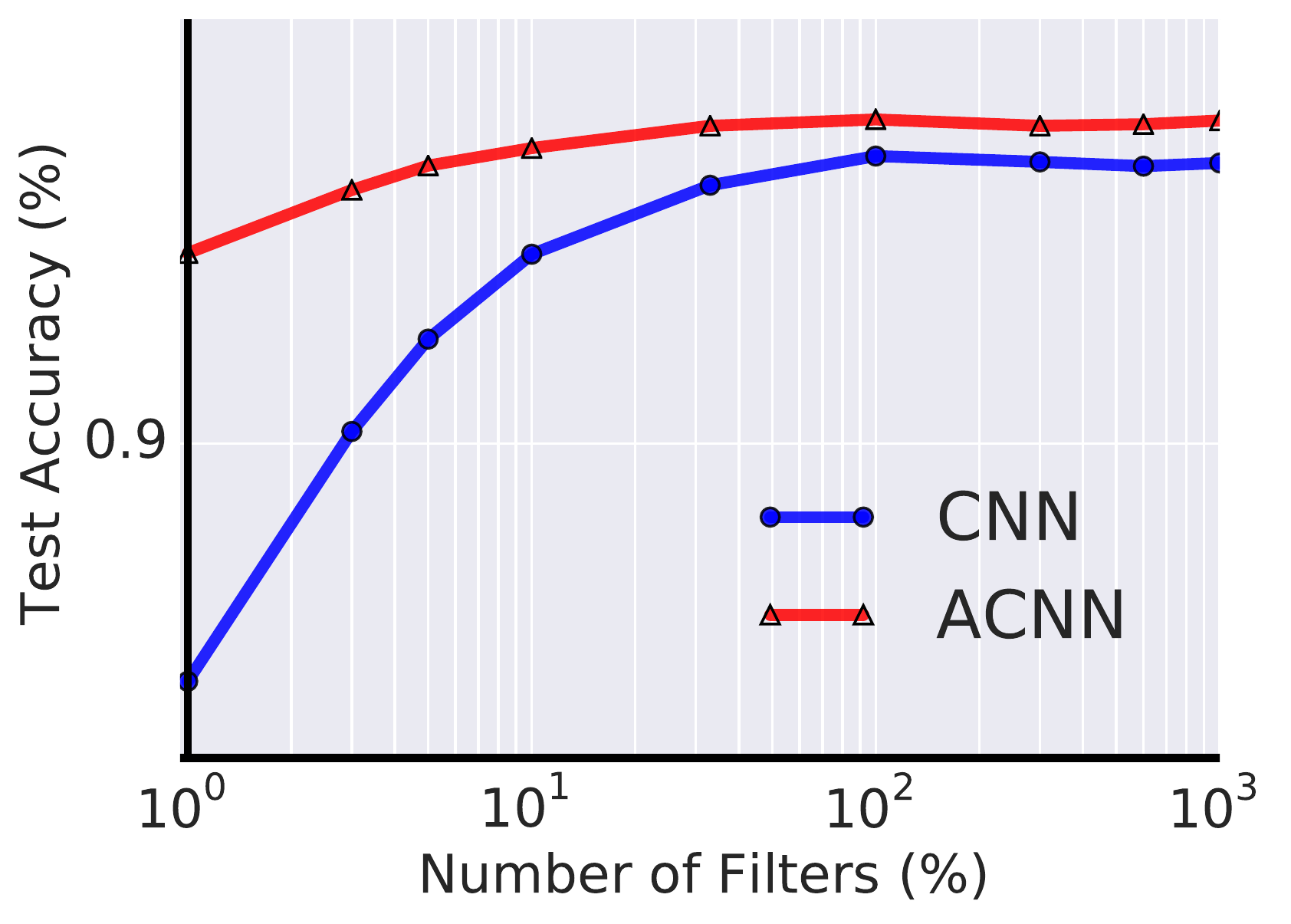} & \hspace{-2mm}
		\includegraphics[width=5.0cm]{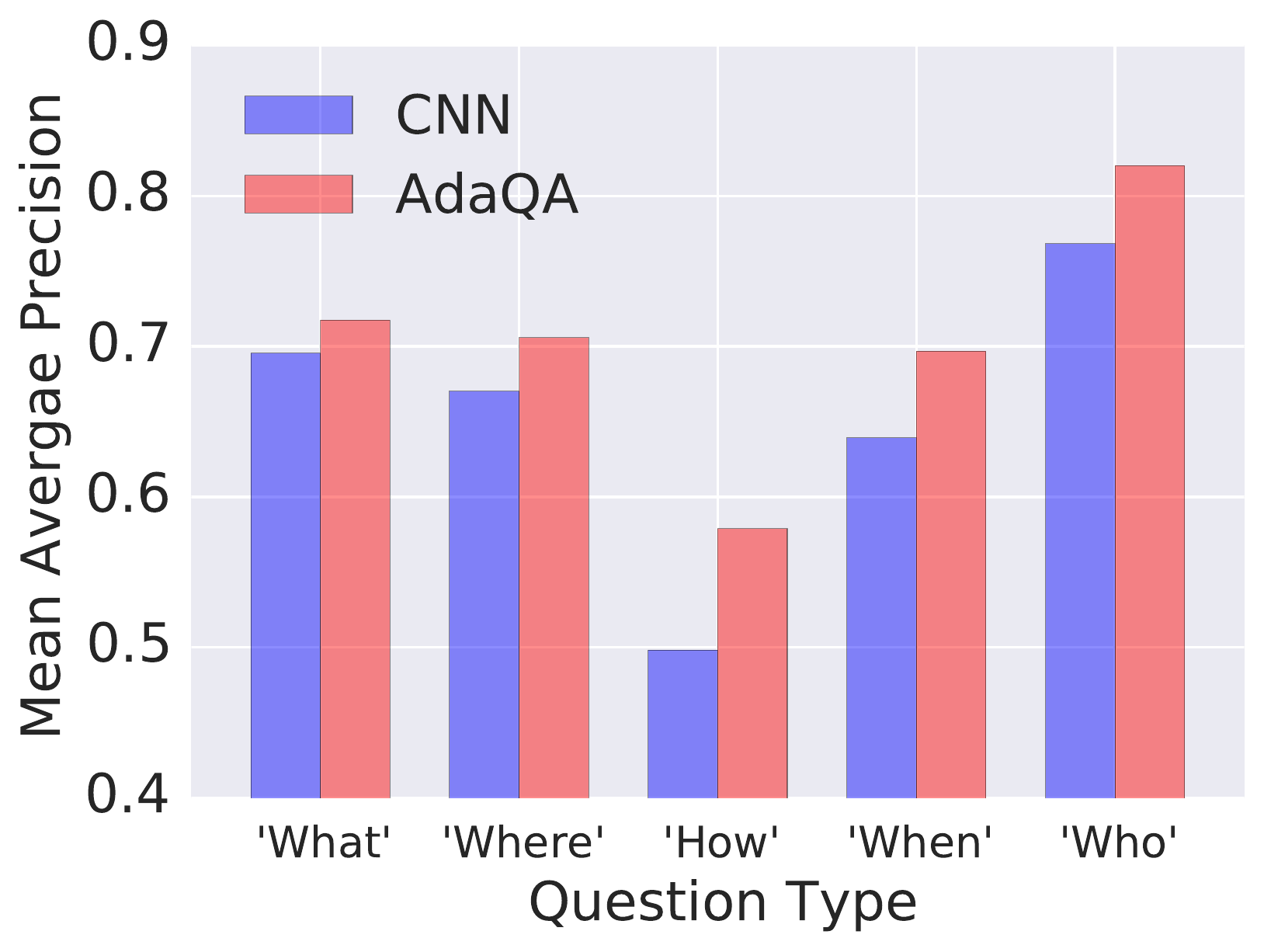} &  \hspace{-2mm}
		\includegraphics[width=5.0cm]{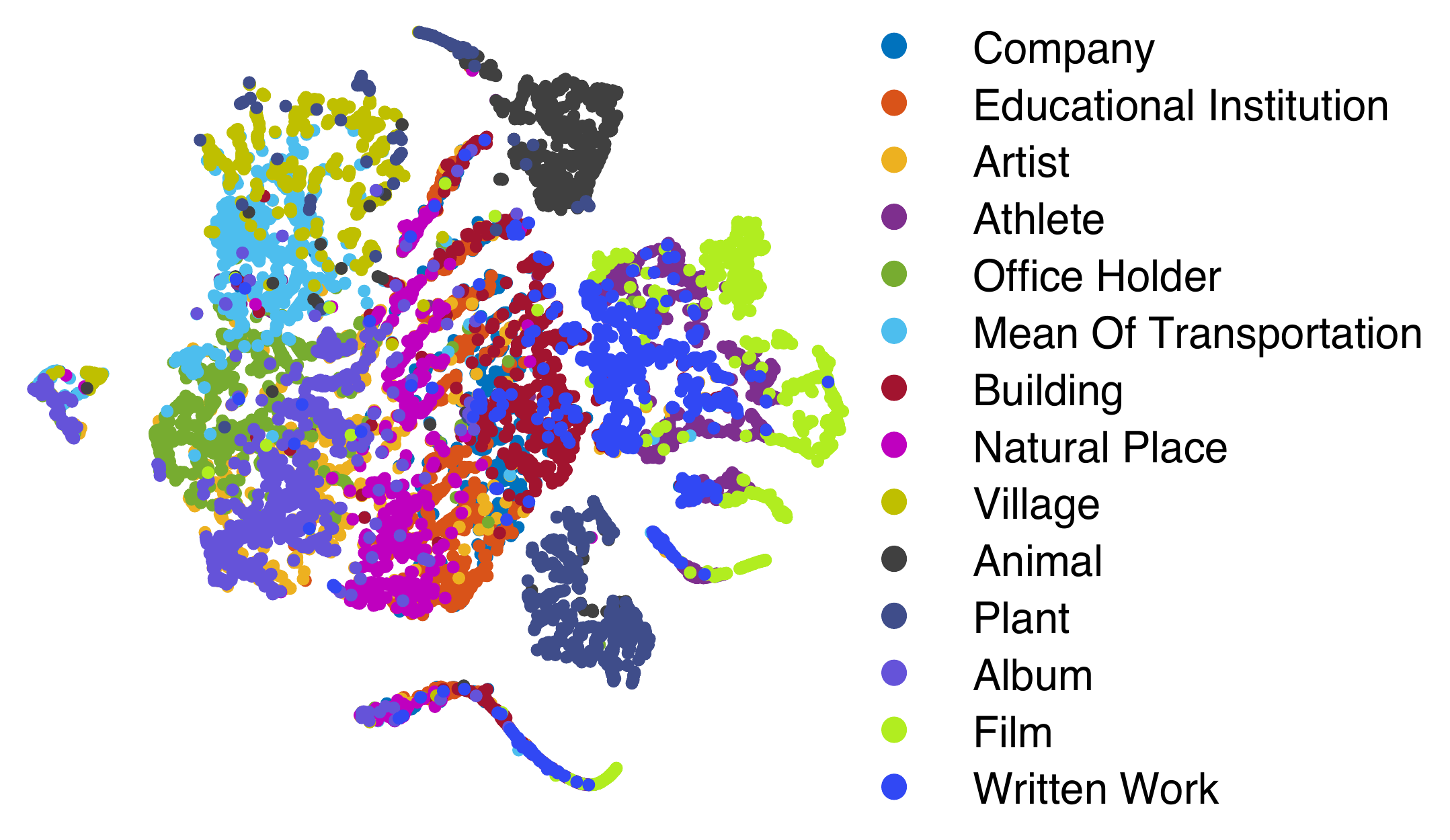} 
		\\
		\hspace{-0mm}
		(a) Effect of filter number \vspace{0mm}  & 
		\hspace{-0mm}
		(b) Different question types \hspace{-0mm}& 
		(c) $t$-SNE visualization  \hspace{-0mm}
	\end{tabular}
	\vspace{-3mm}
	\caption{\small Comprehensive study of the proposed ACNN framework, including (a) the number of filters (Yelp dataset), and (b) performance \emph{vs} question types (WikiQA dataset), and (c) $t$-SNE visualization of learned filter weights (DBpedia dataset).}
	\vspace{-4mm}
	\label{fig:study}
\end{figure*}

We further experiment on both ACNN and CNN models with \emph{multiple} filters. The corresponding document categorization accuracies are presented in Table~\ref{tab:topic}. Although we only use one convolution layer for our ACNN model, it already outperforms other CNN baseline methods with much deeper architectures. Moreover, our method exhibits higher accuracy than n-grams, which is a very strong baseline as shown in \cite{zhang2015character}. We attribute the superior performance of the ACNN framework to its stronger (adaptive) feature-extraction ability. Moreover, our M-ACNN also achieves slightly better performance than self-attentive sentence embeddings proposed in \citet{lin2017structured}, which requires significant more parameters than our method.
%
\paragraph{Effect of number of filters} To further demonstrate that the performance gains in document categorization experiments originates from the improved adaptivity of our ACNN framework, we implement the basic CNN model with different numbers of filter sizes, ranging from $1$ to $1000$. As illustrated in Figure~\ref{fig:study}(a), when the filter size is larger than $100$, the test accuracy of the standard CNN model does not show any noticeable improvement with more filters. More importantly, even with a filter size of $1000$, the classification accuracy of the CNN is worse than that of the ACNN model with the filter number restricted to $100$. Given these observations, we believe that the boosted categorization accuracy does come from the improved flexibility and thus better feature extraction of our ACNN framework.
\begin{table}
	\centering
	\begin{tabular}{c|c|cc}
		\toprule[1.2pt]
		\textbf{Model} &  \textbf{MAP} &  \textbf{MRR}   \\
		\hline
		\multicolumn{3}{c}{\emph{\textbf{CNN-based Baseline Models}}} \\
		\hline
		bigram CNN + \emph{Cnt}$^{\ast}$      & 0.6520 & 0.6652 \\
		Attentive Pooling Network     & 0.6886 & 0.6957 \\
		ABCNN     & 0.6921 & 0.7127 \\
		\hline
		\multicolumn{3}{c}{\emph{\textbf{Our Implementations}}} \\
		\hline
		CNN        & 0.6752 & 0.6890 \\
		ACNN (self-adaptive)         & 0.6897 & 0.7032  \\
		AdaQA (one-way)         & 0.7005 & 0.7161  \\ 
		AdaQA (two-way)       & 0.7107 & 0.7304  \\ 
		AdaQA (two-way) + att. & \textbf{0.7325} & \textbf{0.7428}  \\ 
		\bottomrule[1.2pt]
	\end{tabular}
	\caption{\small Results of our models on WikiQA dataset, compared with previous CNN-based methods.}
	\label{tab:wikiqa}
	\vspace{-5mm}
\end{table}
\vspace{-1mm}
\subsection{Answer Sentence Selection}
\label{sec:exp_qa}
\vspace{-1.5mm}
To elucidate the role of different parts (modules) in our AdaQA model, we implement several model variants for comparison: (\emph{\romannumeral1}) a ``vanilla'' CNN model that independently encodes two sentence representations for matching; (\emph{\romannumeral2}) a self-adaptive ACNN-based model where the question/answer sentence generates adaptive filters only to convolve with the input itself; (\emph{\romannumeral3}) a one-way ACNN model where only the answer sentence representation is extracted with adaptive filters, which are generated conditioned on the question; (\emph{\romannumeral4}) a two-way AdaQA model as described in Section 2.4, where both sentences are adaptively encoded, with filters generated conditioned on the other sequence; (\emph{\romannumeral5}) considering that the proposed filter generation mechanism is complementary to the attention layer typically employed in sequence matching tasks (see Section~\ref{sec:att}), we experiment with another model variant that combines the proposed \emph{context-sensitive} filter generation mechanism with the multi-perspective attention layer introduced in \cite{wang2017bilateral}. 

\begin{table}
	\centering
	\begin{tabular}{c|c|cc}
		\toprule[1.2pt]
		\textbf{Model} &  \textbf{MAP} &  \textbf{MRR}   \\
		\hline
		\multicolumn{3}{c}{\emph{\textbf{CNN-based Baseline Models}}} \\
		\hline
		CNN: \emph{baseline}$^{\ast}$     & 0.8320 &  0.8420 \\
		CNN: \emph{average + word}$^{\ast}$      & 0.8400 & 0.8494 \\
		CNN: \emph{aver + emb}$^{\ast}$     & 0.8466 & 0.8568 \\
		CNN: \emph{hinge\_loss}$^{\ddagger}$     & 0.8758 & 0.8812 \\
		CNN-DAN$^{\ddagger}$    & 0.8655 & 0.8730 \\
		\hline
		\multicolumn{3}{c}{\emph{\textbf{Our Implementations}}} \\
		\hline
		CNN        & 0.8644 & 0.8720 \\
		ACNN (self-adaptive)         & 0.8739 & 0.8801  \\
		AdaQA (one-way)         & 0.8823 & 0.8889  \\ 
		AdaQA (two-way)       & 0.8914 & 0.8983  \\ 
		AdaQA (two-way) + att. & \textbf{0.9021} & \textbf{0.9103}  \\ 
		\bottomrule[1.2pt]
	\end{tabular}
	\caption{\small Results of our models on SelQA dataset, compared with previous CNN-based methods. Results marked with $\ast$ are from \cite{jurczyk2016selqa}, and marked with $\ddagger$ are from \cite{santos2017learning}.}
	\label{tab:selqa}
	\vspace{-6mm}
\end{table}

Tables~\ref{tab:wikiqa} and \ref{tab:selqa} show experimental results of our models on WikiQA and SelQA datasets, along with other state-of-the-art methods. Note that the self-adaptive ACNN model variant, which generates filters only for the input itself (without any interactions before the top matching module), slightly outperforms the vanilla CNN Siamese model. Combined with the results in document categorization experiments, we believe that our ACNN framework, in its simplest form, can be utilized as a powerful feature extractor for transforming natural language sentences into fixed-length vectors. 
More importantly, our two-way AdaQA model exhibits superior results compared with the one-way variant as well as other CNN-based baseline models on the WikiQA dataset. This observation indicates that the \emph{bidirectional} filter generation mechanism is strongly associated with the performance gains. 
While combined with the multi-perspective attention layers, adopted after the ACNN encoding layer, our two-way AdaQA model achieves even better performance. This suggests that the proposed strategy is complementary, in terms of the incorporation of rich contextual information, to the standard attention mechanism.  The same trend is also observed on the SelQA dataset (as shown in Table~\ref{tab:selqa}), which is a much larger dataset than WikiQA. 

Notably, our model yields significantly better results than an attentive pooling network and ABCNN (attention-based CNN) baselines. We attribute the improvement to two potential advantages of our AdaQA model: (\emph{\romannumeral1}) for the two previous baseline methods, the interaction between question and answer takes place either before or after convolution. However, in our AdaQA model, the communication between two sentences is inherent in the convolution operation, and thus can provide the abstracted features with more flexibility; (\emph{\romannumeral2}) the \emph{bidirectional} filter generation mechanism in our AdaQA model generates co-dependent representations for the question and candidate answer, which could enable the model to recover from initial local maxima corresponding to incorrect predictions \cite{xiong2016dynamic}. 
%

\begin{table} \centering
	\begin{tabular}{c|c}
		\toprule[1.2pt]
		\textbf{Model} &  \textbf{Accuracy}    \\
		\hline
		Siamese-CNN      & 0.7960 \\
		Multi-Perspective-CNN    & 0. 8138 \\
		\hline
		AdaQA (two-way)  & 0.8516   \\ 
		AdaQA (two-way) + att. & \textbf{0.8794}   \\ 
		\bottomrule[1.2pt]
	\end{tabular}
	\vspace{-2mm}
	\caption{\small Results on the Quora Question Pairs dataset.}
	\label{tab:quora}
	\vspace{-5mm}
\end{table}
\vspace{-2mm}
\paragraph{Paragraph Identification} Considering that the proposed AdaQA model can be readily generalized to other text sequence matching problems, we further evaluate the proposed framework on the paraphrase identification task with the Quora question pairs dataset. To ensure a fair comparison, we employ the same data splits as in \cite{wang2017bilateral}. As illustrated in Table~\ref{tab:quora}, our two-way AdaQA model  again exhibits superior performances compared with basic CNN models (as reported in \cite{wang2017bilateral}).
%

\subsection{Discussion}
\vspace{-1mm}
\paragraph{Reasoning ability}
To associate the improved answer sentence selection results with the reasoning capabilities of our AdaQA model, we further categorize the questions in the WikiQA test set into 5 types containing: `What', `Where', `How', `When' or `Who'. We then calculate the MAP scores of the basic CNN and our AdaQA model on different question types. Similar to the findings in \cite{miao2016neural}, we observe that the `How' question is the hardest to answer, with the lowest MAP scores. However, our AdaQA model improves most over the basic CNN on the `How' type question, see Figure~\ref{fig:study}(b). Further comparing our results with NASM in \cite{miao2016neural}, our AdaQA model (with a MAP score of $0.579$) outperforms their reported `How' question MAP scores ($0.524$) by a large margin, indicating that the adaptive convolutional filter-generation mechanism improves the model's ability to read and reason over natural language sentences.
\vspace{-2mm}
\paragraph{Filter visualization}
To better understand what information has been encoded into our \emph{context-sensitive} filters, we visualize one of the filters for sentences within the test set  (on the DBpedia dataset) with $t$-SNE. The corresponding results are shown in Figure~\ref{fig:study}(c). It can be observed that the filters for documents with the same label (ontology) are grouped into clusters, indicating that for different types of document, ACNN has leveraged distinct convolutional filters for  better feature extraction.
\vspace{-1mm}
\section{Conclusions}
\vspace{-2mm}
We presented a \emph{context-sensitive} convolutional filter-generation mechanism, introducing a meta network to adaptively produce a set of input-aware filters. In this manner, the filter weights vary from sample to sample, providing the CNN encoder network with more modeling flexibility and capacity. This framework is further generalized to model question-answer sentence pairs, leveraging a two-way feature abstraction process. We evaluate our models on several document-categorization and sentence matching benchmarks, and they consistently outperform the standard CNN and attention-based CNN baselines, demonstrating the effectiveness of our framework.

\paragraph{Acknowledgments}
This research was supported in part by DARPA, DOE, NIH, ONR and NSF.

\bibliography{acl2018}

\begin{thebibliography}{43}
\expandafter\ifx\csname natexlab\endcsname\relax\def\natexlab#1{#1}\fi

\bibitem[{Abadi et~al.(2016)Abadi, Barham, Chen, Chen, Davis, Dean, Devin,
  Ghemawat, Irving, Isard et~al.}]{abadi2016tensorflow}
Mart{\'\i}n Abadi, Paul Barham, Jianmin Chen, Zhifeng Chen, Andy Davis, Jeffrey
  Dean, Matthieu Devin, Sanjay Ghemawat, Geoffrey Irving, Michael Isard, et~al.
  2016.
\newblock Tensorflow: A system for large-scale machine learning.
\newblock In \emph{OSDI}, volume~16, pages 265--283.

\bibitem[{Bahdanau et~al.(2015)Bahdanau, Cho, and Bengio}]{bahdanau2014neural}
Dzmitry Bahdanau, Kyunghyun Cho, and Yoshua Bengio. 2015.
\newblock Neural machine translation by jointly learning to align and
  translate.
\newblock \emph{ICLR}.

\bibitem[{Bromley et~al.(1994)Bromley, Guyon, LeCun, S{\"a}ckinger, and
  Shah}]{bromley1994signature}
Jane Bromley, Isabelle Guyon, Yann LeCun, Eduard S{\"a}ckinger, and Roopak
  Shah. 1994.
\newblock Signature verification using a" siamese" time delay neural network.
\newblock In \emph{NIPS}, pages 737--744.

\bibitem[{Cohen and Singer(1999)}]{cohen1999context}
William~W Cohen and Yoram Singer. 1999.
\newblock Context-sensitive learning methods for text categorization.
\newblock \emph{TOIS}, 17(2):141--173.

\bibitem[{Collobert et~al.(2011)Collobert, Weston, Bottou, Karlen, Kavukcuoglu,
  and Kuksa}]{collobert2011natural}
Ronan Collobert, Jason Weston, L{\'e}on Bottou, Michael Karlen, Koray
  Kavukcuoglu, and Pavel Kuksa. 2011.
\newblock Natural language processing (almost) from scratch.
\newblock \emph{JMLR}, 12(Aug):2493--2537.

\bibitem[{Conneau et~al.(2016)Conneau, Schwenk, Barrault, and
  Lecun}]{conneau2016very}
Alexis Conneau, Holger Schwenk, Lo{\"\i}c Barrault, and Yann Lecun. 2016.
\newblock Very deep convolutional networks for text classification.
\newblock \emph{EACL}.

\bibitem[{Dauphin et~al.(2017)Dauphin, Fan, Auli, and
  Grangier}]{dauphin2016language}
Yann~N Dauphin, Angela Fan, Michael Auli, and David Grangier. 2017.
\newblock Language modeling with gated convolutional networks.
\newblock \emph{ICML}.

\bibitem[{De~Brabandere et~al.(2016)De~Brabandere, Jia, Tuytelaars, and
  Van~Gool}]{de2016dynamic}
Bert De~Brabandere, Xu~Jia, Tinne Tuytelaars, and Luc Van~Gool. 2016.
\newblock Dynamic filter networks.
\newblock In \emph{NIPS}.

\bibitem[{Gan et~al.(2016)Gan, Pu, Henao, Li, He, and Carin}]{gan2016learning}
Zhe Gan, Yunchen Pu, Ricardo Henao, Chunyuan Li, Xiaodong He, and Lawrence
  Carin. 2016.
\newblock Learning generic sentence representations using convolutional neural
  networks.
\newblock \emph{arXiv preprint arXiv:1611.07897}.

\bibitem[{Gehring et~al.(2017)Gehring, Auli, Grangier, Yarats, and
  Dauphin}]{gehring2017convolutional}
Jonas Gehring, Michael Auli, David Grangier, Denis Yarats, and Yann~N Dauphin.
  2017.
\newblock Convolutional sequence to sequence learning.
\newblock \emph{ICML}.

\bibitem[{Ha et~al.(2017)Ha, Dai, and Le}]{ha2016hypernetworks}
David Ha, Andrew Dai, and Quoc~V Le. 2017.
\newblock Hypernetworks.
\newblock \emph{ICLR}.

\bibitem[{Hu et~al.(2014)Hu, Lu, Li, and Chen}]{hu2014convolutional}
Baotian Hu, Zhengdong Lu, Hang Li, and Qingcai Chen. 2014.
\newblock Convolutional neural network architectures for matching natural
  language sentences.
\newblock In \emph{NIPS}, pages 2042--2050.

\bibitem[{Jeon and Kim(2017)}]{jeon2017active}
Yunho Jeon and Junmo Kim. 2017.
\newblock Active convolution: Learning the shape of convolution for image
  classification.
\newblock \emph{CVPR}.

\bibitem[{Jurczyk et~al.(2016)Jurczyk, Zhai, and Choi}]{jurczyk2016selqa}
Tomasz Jurczyk, Michael Zhai, and Jinho~D Choi. 2016.
\newblock Selqa: A new benchmark for selection-based question answering.
\newblock In \emph{ICTAI}, pages 820--827. IEEE.

\bibitem[{Kim(2014)}]{kim2014convolutional}
Yoon Kim. 2014.
\newblock Convolutional neural networks for sentence classification.
\newblock \emph{EMNLP}.

\bibitem[{Kingma and Ba(2014)}]{kingma2014adam}
Diederik Kingma and Jimmy Ba. 2014.
\newblock Adam: A method for stochastic optimization.
\newblock \emph{arXiv preprint arXiv:1412.6980}.

\bibitem[{LeCun et~al.(1998)LeCun, Bottou, Bengio, and
  Haffner}]{lecun1998gradient}
Yann LeCun, L{\'e}on Bottou, Yoshua Bengio, and Patrick Haffner. 1998.
\newblock Gradient-based learning applied to document recognition.
\newblock \emph{Proceedings of the IEEE}, 86(11):2278--2324.

\bibitem[{Lin et~al.(2017)Lin, Feng, Santos, Yu, Xiang, Zhou, and
  Bengio}]{lin2017structured}
Zhouhan Lin, Minwei Feng, Cicero Nogueira~dos Santos, Mo~Yu, Bing Xiang, Bowen
  Zhou, and Yoshua Bengio. 2017.
\newblock A structured self-attentive sentence embedding.
\newblock \emph{ICLR}.

\bibitem[{Liu et~al.(2017)Liu, Qiu, and Huang}]{liu2017dynamic}
Pengfei Liu, Xipeng Qiu, and Xuanjing Huang. 2017.
\newblock Dynamic compositional neural networks over tree structure.
\newblock \emph{IJCAI}.

\bibitem[{Miao et~al.(2016)Miao, Yu, and Blunsom}]{miao2016neural}
Yishu Miao, Lei Yu, and Phil Blunsom. 2016.
\newblock Neural variational inference for text processing.
\newblock In \emph{ICML}, pages 1727--1736.

\bibitem[{Mikolov et~al.(2013)Mikolov, Sutskever, Chen, Corrado, and
  Dean}]{mikolov2013distributed}
Tomas Mikolov, Ilya Sutskever, Kai Chen, Greg~S Corrado, and Jeff Dean. 2013.
\newblock Distributed representations of words and phrases and their
  compositionality.
\newblock In \emph{NIPS}, pages 3111--3119.

\bibitem[{Mou et~al.(2016)Mou, Men, Li, Xu, Zhang, Yan, and
  Jin}]{mou2015natural}
Lili Mou, Rui Men, Ge~Li, Yan Xu, Lu~Zhang, Rui Yan, and Zhi Jin. 2016.
\newblock Natural language inference by tree-based convolution and heuristic
  matching.
\newblock \emph{ACL}.

\bibitem[{Pennington et~al.(2014)Pennington, Socher, and
  Manning}]{pennington2014glove}
Jeffrey Pennington, Richard Socher, and Christopher~D. Manning. 2014.
\newblock Glove: Global vectors for word representation.
\newblock In \emph{EMNLP}, pages 1532--1543.

\bibitem[{Radford et~al.(2016)Radford, Metz, and
  Chintala}]{radford2015unsupervised}
Alec Radford, Luke Metz, and Soumith Chintala. 2016.
\newblock Unsupervised representation learning with deep convolutional
  generative adversarial networks.
\newblock \emph{ICLR}.

\bibitem[{dos Santos et~al.(2016)dos Santos, Tan, Xiang, and
  Zhou}]{dos2016attentive}
C{\i}cero~Nogueira dos Santos, Ming Tan, Bing Xiang, and Bowen Zhou. 2016.
\newblock Attentive pooling networks.
\newblock \emph{CoRR, abs/1602.03609}.

\bibitem[{Santos et~al.(2017)Santos, Wadhawan, and Zhou}]{santos2017learning}
Cicero Nogueira~dos Santos, Kahini Wadhawan, and Bowen Zhou. 2017.
\newblock Learning loss functions for semi-supervised learning via
  discriminative adversarial networks.
\newblock \emph{arXiv preprint arXiv:1707.02198}.

\bibitem[{Shen et~al.(2018{\natexlab{a}})Shen, Su, Chapfuwa, Wang, Wang, Carin,
  and Henao}]{Shen2018NASHTE}
Dinghan Shen, Qinliang Su, Paidamoyo Chapfuwa, Wenlin Wang, Guoyin Wang,
  Lawrence Carin, and Ricardo Henao. 2018{\natexlab{a}}.
\newblock Nash: Toward end-to-end neural architecture for generative semantic
  hashing.
\newblock In \emph{ACL}.

\bibitem[{Shen et~al.(2018{\natexlab{b}})Shen, Wang, Wang, Min, Su, Zhang, Li,
  Henao, and Carin}]{Shen2018BaselineNM}
Dinghan Shen, Guoyin Wang, Wenlin Wang, Martin~Renqiang Min, Qinliang Su, Yizhe
  Zhang, Chunyuan Li, Ricardo Henao, and Lawrence Carin. 2018{\natexlab{b}}.
\newblock Baseline needs more love: On simple word-embedding-based models and
  associated pooling mechanisms.
\newblock In \emph{ACL}.

\bibitem[{Shen et~al.(2017)Shen, Zhang, Henao, Su, and
  Carin}]{shen2017deconvolutional}
Dinghan Shen, Yizhe Zhang, Ricardo Henao, Qinliang Su, and Lawrence Carin.
  2017.
\newblock Deconvolutional latent-variable model for text sequence matching.
\newblock \emph{arXiv preprint arXiv:1709.07109}.

\bibitem[{Srivastava et~al.(2014)Srivastava, Hinton, Krizhevsky, Sutskever, and
  Salakhutdinov}]{srivastava2014dropout}
Nitish Srivastava, Geoffrey~E Hinton, Alex Krizhevsky, Ilya Sutskever, and
  Ruslan Salakhutdinov. 2014.
\newblock Dropout: a simple way to prevent neural networks from overfitting.
\newblock \emph{JMLR}, 15(1):1929--1958.

\bibitem[{Tang et~al.(2018)Tang, Jin, Fang, Wang, and Sa}]{tang2018speeding}
Shuai Tang, Hailin Jin, Chen Fang, Zhaowen Wang, and Virginia Sa. 2018.
\newblock Speeding up context-based sentence representation learning with
  non-autoregressive convolutional decoding.
\newblock In \emph{Proceedings of The Third Workshop on Representation Learning
  for NLP}, pages 69--78.

\bibitem[{Tang and de~Sa(2018)}]{tang2018multi}
Shuai Tang and Virginia~R de~Sa. 2018.
\newblock Multi-view sentence representation learning.
\newblock \emph{arXiv preprint arXiv:1805.07443}.

\bibitem[{Thrun and Pratt(2012)}]{thrun2012learning}
Sebastian Thrun and Lorien Pratt. 2012.
\newblock \emph{Learning to learn}.
\newblock Springer Science \& Business Media.

\bibitem[{Wang et~al.(2018)Wang, Li, Wang, Zhang, Shen, Zhang, Henao, and
  Carin}]{wang2018joint}
Guoyin Wang, Chunyuan Li, Wenlin Wang, Yizhe Zhang, Dinghan Shen, Xinyuan
  Zhang, Ricardo Henao, and Lawrence Carin. 2018.
\newblock Joint embedding of words and labels for text classification.
\newblock \emph{arXiv preprint arXiv:1805.04174}.

\bibitem[{Wang and Jiang(2017)}]{wang2016compare}
Shuohang Wang and Jing Jiang. 2017.
\newblock A compare-aggregate model for matching text sequences.
\newblock \emph{ICLR}.

\bibitem[{Wang et~al.(2017{\natexlab{a}})Wang, Gan, Wang, Shen, Huang, Ping,
  Satheesh, and Carin}]{wang2017topic}
Wenlin Wang, Zhe Gan, Wenqi Wang, Dinghan Shen, Jiaji Huang, Wei Ping, Sanjeev
  Satheesh, and Lawrence Carin. 2017{\natexlab{a}}.
\newblock Topic compositional neural language model.
\newblock \emph{arXiv preprint arXiv:1712.09783}.

\bibitem[{Wang et~al.(2017{\natexlab{b}})Wang, Hamza, and
  Florian}]{wang2017bilateral}
Zhiguo Wang, Wael Hamza, and Radu Florian. 2017{\natexlab{b}}.
\newblock Bilateral multi-perspective matching for natural language sentences.
\newblock \emph{IJCAI}.

\bibitem[{Wang et~al.(2016)Wang, Mi, and Ittycheriah}]{wang2016sentence}
Zhiguo Wang, Haitao Mi, and Abraham Ittycheriah. 2016.
\newblock Sentence similarity learning by lexical decomposition and
  composition.
\newblock \emph{COLING}.

\bibitem[{Xiong et~al.(2017)Xiong, Zhong, and Socher}]{xiong2016dynamic}
Caiming Xiong, Victor Zhong, and Richard Socher. 2017.
\newblock Dynamic coattention networks for question answering.
\newblock \emph{ICLR}.

\bibitem[{Yang et~al.(2015)Yang, Yih, and Meek}]{yang2015wikiqa}
Yi~Yang, Wen-tau Yih, and Christopher Meek. 2015.
\newblock Wikiqa: A challenge dataset for open-domain question answering.
\newblock In \emph{EMNLP}, pages 2013--2018.

\bibitem[{Yin et~al.(2016)Yin, Sch{\"u}tze, Xiang, and Zhou}]{yin2015abcnn}
Wenpeng Yin, Hinrich Sch{\"u}tze, Bing Xiang, and Bowen Zhou. 2016.
\newblock Abcnn: Attention-based convolutional neural network for modeling
  sentence pairs.
\newblock \emph{TACL}.

\bibitem[{Zhang et~al.(2015)Zhang, Zhao, and LeCun}]{zhang2015character}
Xiang Zhang, Junbo Zhao, and Yann LeCun. 2015.
\newblock Character-level convolutional networks for text classification.
\newblock In \emph{NIPS}, pages 649--657.

\bibitem[{Zhang et~al.(2018)Zhang, Li, Shen, and Carin}]{zhang2018diffusion}
Xinyuan Zhang, Yitong Li, Dinghan Shen, and Lawrence Carin. 2018.
\newblock Diffusion maps for textual network embedding.
\newblock \emph{arXiv preprint arXiv:1805.09906}.

\end{thebibliography}
\bibliographystyle{acl_natbib_nourl}

\appendix


\end{document}